\documentclass{article}

\PassOptionsToPackage{numbers, compress}{natbib}

    \usepackage[preprint]{neurips_2024}

\usepackage[utf8]{inputenc} 
\usepackage[T1]{fontenc}    
\usepackage{hyperref}       
\usepackage{url}            
\usepackage{booktabs}       
\usepackage{amsfonts}       
\usepackage{nicefrac}       
\usepackage{microtype}      
\usepackage{color}         
\usepackage[table]{xcolor}
\usepackage{graphicx}
\usepackage{amsmath} 
\usepackage{multirow} 
\usepackage{cite} 
\graphicspath{{./figures/}}

\title{Rethinking 3D Dense Caption and Visual Grounding in A Unified Framework through Prompt-based Localization}

\author{%
  Yongdong Luo\thanks{Equal contribution} \\
  Xiamen University
  \And
  Haojia Lin\footnotemark[\value{footnote}] \\
  Xiamen University 
  \And
  Xiawu Zheng \\
  Xiamen University 
  \And
  Yigeng Jiang \\
  Xiamen University 
  \AND
  Fei Chao \\
  Xiamen University 
  \And
  Jie Hu 
  \And
  Guannan Jiang 
  \And
  Songan Zhang 
  \And
  Rongrong Ji \\
  Xiamen University
}

\begin{document}

\maketitle

\begin{abstract}
3D Visual Grounding (3DVG) and 3D Dense Captioning (3DDC) are two crucial tasks in various 3D applications, which require both shared and complementary information in localization and visual-language relationships. 
Therefore, existing approaches adopt the two-stage ``detect-then-describe/discriminate" pipeline, which relies heavily on the performance of the detector, resulting in suboptimal performance.
Inspired by DETR, we propose a unified framework, 3DGCTR, to jointly solve these two distinct but closely related tasks in an end-to-end fashion. 
The key idea is to reconsider the prompt-localization ability of the 3DVG model. 
In this way, the 3DVG model with a well-designed prompt as input can assist the 3DDC task by extracting localization information from the prompt. 
In terms of implementation, we integrate a caption head into the existing 3DVG network with a caption text prompt as a connection, effectively harnessing the existing 3DVG model's inherent localization capacity, thereby boosting 3DDC capability. This integration facilitates simultaneous multi-task training on both tasks, mutually enhancing their performance.
Extensive experimental results demonstrate the effectiveness of this approach. Specifically, on the ScanRefer dataset, 3DGCTR surpasses the state-of-the-art 3DDC method by 4.30\% in CIDEr@0.5IoU in MLE training and improves upon the SOTA 3DVG method by 3.16\% in Acc@0.25IoU. The codes are at \url{https://github.com/Leon1207/3DGCTR}.
\end{abstract}

\section{Introduction}\label{intro}
The intersection of 3D scene understanding and natural language processing has emerged as a focal point in research, evident in tasks like 3D visual grounding (3DVG) \citep{viewrefer,multi3drefer,ns3d,ham,butd,eda,3dvg,sat,instancerefer,languagerefer,3dsps,lamm,tgnn} and 3D dense captioning (3DDC) \citep{Xtrans2cap,scan2cap,more,contextual,3ddc_survey}. 
3DVG involves processing a text paired with a point cloud to identify the specified object with a bounding box.
Its counterpart, 3DDC, takes a point cloud and produces detailed bounding boxes and descriptions for each object.
These tasks are pivotal for developing innovative applications, including assistive robotics and intuitive language-based interaction in AR/VR environments.

Considering that 3DVG and 3DDC contain both shared and complementary information in nature, previous works \citep{d3net,3djcg,unit3d} attempt to integrate these two tasks into a unified work. 
D3Net \citep{d3net} proposes a speaker-listener pipeline to improve the performance of these two tasks in a self-critical manner.
Based on a pre-trained detector, 3DJCG \citep{3djcg} utilizes an attribute and relation-aware module to enhance the task-agnostic features. 
Unit3D \citep{unit3d} propose a transformer-based fusion module that is based on a PointGroup \citep{pointgroup} detection backbone and a BERT \citep{bert} encoder, to learn the multi-model representations between objects and text inputs. 
Among existing methods, they all adopt a detector-based architecture \citep{d3net,3djcg,unit3d}, in which two task-specific modules were stacked on a shared detector \citep{votenet,pointgroup} to unify these two tasks in a single framework. Though these \textit{two-stage} methods have achieved remarkable performance, the detector-based architecture yields suboptimal performance due to the following issues:
\textbf{1) For 3DVG}, the language-irrelevant proposals generated by the detector easily cause incorrect localization, which is demonstrated by 3D-SPS \citep{3dsps}. 
\textbf{2) For 3DDC}, the serial and explicit reasoning highly limits the mutual promotion of localization and captioning.

\begin{figure}[t!]
  \centering 
  \includegraphics[width=1.0\linewidth]{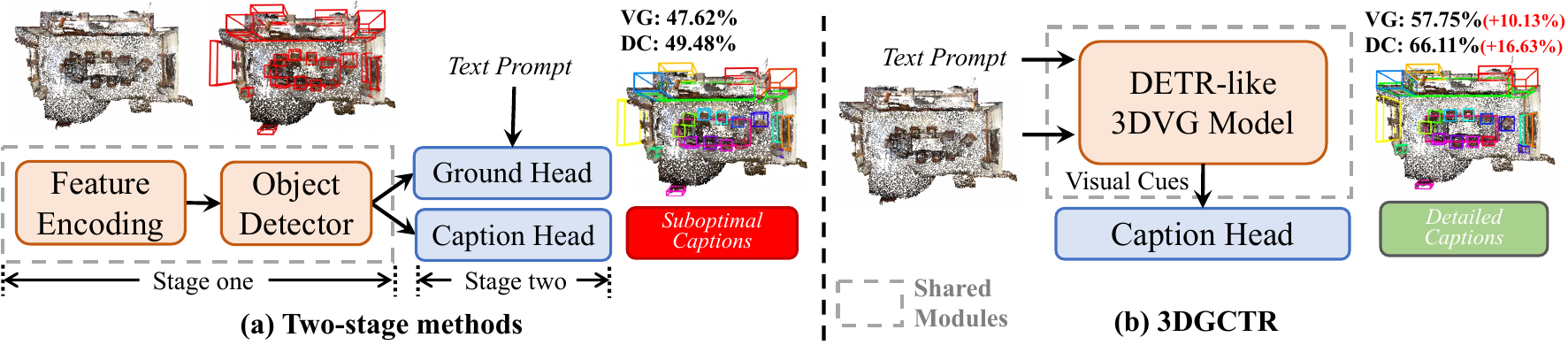}
  \caption{Illustration of existing method (a) with two-stage pipeline and our single-stage 3DGCTR (b). Existing methods heavily depend on a detector’s output and also suffer from low reuse of task-agnostic modules. Therefore, we propose a transformer-based model that simply builds upon a mature 3DVG model, thus giving 3DVG model 3DDC capability. Compared to the SOTA method 3DJCG \citep{3djcg} that jointly trains the two tasks, our method achieves a significant improvement.}
  \label{fig_intro} 
\end{figure}

Recently, DETR-like \citep{detr} models developed based on Transformers \citep{transformer} have achieved inspiring progress on both 3DVG \citep{3dsps,butd,eda} and 3DDC \citep{vote2cap} tasks in a \textit{single-stage} way. 
It is a natural idea to integrate these two tasks into a single DETR-like architecture, thus maximizing the use of task-agnostic modules to achieve end-to-end training.
However, this idea is not easy to implement due to the following reasons: 
\textbf{1) Variations in Input Types.} In terms of inputs, 3DDC exclusively requires a point cloud from a scene, whereas 3DVG also demands additional text references. 
\textbf{2) Divergent Optimization Objectives for Query Embeddings.} Query embeddings are an important component of the DETR-like architecture, which represents the object proposals to be located in a scene. In 3DVG methods \citep{3dsps,butd,eda}, the query embeddings are learned to align the target object described in the input text. In contrast, 3DDC \citep{vote2cap} focuses on aligning query embeddings with all predetermined ground-truth objects in the scene.

To address the above issues and naturally merge two different tasks, we rethink the role of a well-crafted grounding referencing text. As described in \citep{butd} and \citep{eda}, they leverage prompt-based detection for data augmentation. However, by utilizing such a prompt as a link, it is possible to cast the detection part of the dense caption model as referential grounding.
Specifically, as described above, the query embeddings are trained to align the input text in the 3DVG task. When a prompt such as  ``\textit{cabinet . bed . chair . sofa . }" composed of multiple object names is fed into the 3DVG model, the 3DVG model can locate the objects mentioned in the prompt, thus turning it into a detector.
By harmonizing the input form and the optimization objectives for query embeddings across both tasks, it's natural to merge the two tasks into the DETR-like architecture. This eliminates the two-stage optimization paradigm, where two tasks share visual representations and are optimized simultaneously during training, reinforcing each other.
With this rethinking, we proposed a unified Transformer framework termed 3DGCTR, which integrates the Dual-Clued Captioner, a lightweight caption head outlined in \citep{vote2cap}, into the 3DVG model. Together with different text prompts for the two tasks, we enable the model to train both tasks end-to-end.

Experiments show that both the 3DDC and 3DVG performance of 3DGCTR have achieved state-of-the-art on the ScanRefer\citep{scanrefer} benchmark. To be specific, 3DGCTR surpasses the 3DDC method by 4.30\% in CIDEr@0.5IoU in MLE training and improves upon the 3DVG method by 3.16\% in Acc@0.25IoU.
Meanwhile, through joint training, 3DGCTR can achieve the mutual promotion of the two tasks. Specifically, the 3DDC and 3DVG tasks increased CIDEr@0.5IoU by 1.27\% and Acc@0.25IoU by 0.3\%, respectively.

To sum up, the main contributions are as follows:

\begin{itemize}
    \item \textbf{Rethinking the role of prompts in the 3DVG model.} We provide a new perspective on the prompt-localization ability of the 3DVG model, transforming it into a dual-purpose tool that effectively facilitates both 3DVG and 3DDC tasks.
    \item \textbf{Advanced Unified Framework for 3D Visual Grounding and 3D Dense Captioning.} By adding a caption head to the 3DVG model and using a well-designed prompt for the 3DDC task, our 3DGCTR framework integrates 3DVG and 3DDC tasks within a DETR-like architecture, enabling efficient end-to-end training.
    \item \textbf{State-of-the-Art Performance.} Not only does our integration approach produce outstanding results, but in our end-to-end training, both DC and VG tasks were mutually enhanced, which set new benchmarks for both 3D captioning and grounding tasks, further establishing our framework's dominance in the domain.
\end{itemize}

\section{Related Work}

\subsection{3D Visual Grounding}

3D Visual Grounding (3DVG) aims to identify the targeted object in a 3D scene based on linguistic cues. 
Broadly, 3DVG encompasses two pivotal tasks: 3DREC and 3DRES. 

\textbf{Two-Stage Method.} Within the 3DREC realm, most recent works \citep{3dvg,sat} employ a two-stage pipeline. Initially, they utilize either ground truth or a 3D object detector \citep{pointgroup,groupfree} for generating object proposals. Subsequently, these methods use text and 3D encoder \citep{roberta,pointnet2} to extract features and then ground the target one after the feature fusion. 
For instance, InstanceRefer \citep{instancerefer} ingeniously construes the task as instance-matching, while LanguageRefer \citep{languagerefer} reinterprets it as a linguistic modeling task by replacing the 3D features with predicted object labels. 

\textbf{Single-Stage Method.} 3D-SPS \citep{3dsps} is the first to introduce a single-stage network for 3DREC. Meanwhile, recent state-of-the-art 3DREC approaches, such as BUTD-DETR \citep{butd} and EDA \citep{eda}, have incorporated the query selection modules from Groupfree \citep{groupfree}, indicating a converging trend of these technologies. It is worth noting that the two-stage setting introduced in \citep{eda} is different from that described in our paper. Specifically, the outputs of the detector participate in the cross-attention of the query embedding as auxiliary information, which in essence still belongs to the single-stage training method. In comparison, the 3DRES field remains largely unexplored, with the only notable single-stage method TGNN \citep{tgnn}. TGNN builds on an instance segmentation framework, employing text-guided Graph Neural Networks to pinpoint the center of the instance referred to in the text.
In this paper, we will focus our research on 3DREC because it has been fully explored.

\subsection{3D Dense Captioning}

3D dense captioning is a task that requires translating 3D scene information to a set of bounding boxes and natural language descriptions.

\textbf{Two-Stage Method.} Scan2Cap \citep{scan2cap} and MORE \citep{more} create graphs based on a detector's \citep{deep_hough, pointgroup} box predictions, using predefined rules to decipher complex object relations in 3D scenes. SpaCap3D \citep{spacap} develops a spatially-informed transformer to understand spatial relationships among detector outputs. 3DJCG \citep{3djcg} and D3Net \citep{d3net} explore the mutual enhancement of 3D dense captioning and 3D visual grounding. $\mathcal{X}$-Trans2Cap \citep{Xtrans2cap} incorporates 2D priors to enhance 3D dense captioning through knowledge transfer. Recently, \citep{zhong2022contextual} has focused on contextual data for recognizing non-object details. These methods have significantly advanced the 3D dense captioning challenge. Yet, they are heavily reliant on the accuracy of the detector.

\textbf{Single-Stage Method.} Vote2Cap-DETR \citep{vote2cap} is a unified, single-stage approach that concurrently detects objects and generates captions, addressing 3D dense captioning as a set prediction issue.
Mirroring Vote2Cap-DETR \citep{vote2cap}, our 3DGCTR model seamlessly integrates 3DDC and 3DVG tasks within a DETR-like framework. Furthermore, 3DGCTR's visual grounding aspect enhances the dense captioning task more effectively than joint training of detection and captioning tasks, owing to its superior text-visual alignment.

\begin{figure*}[t!]
  \centering 
  \includegraphics[width=1.0\linewidth]{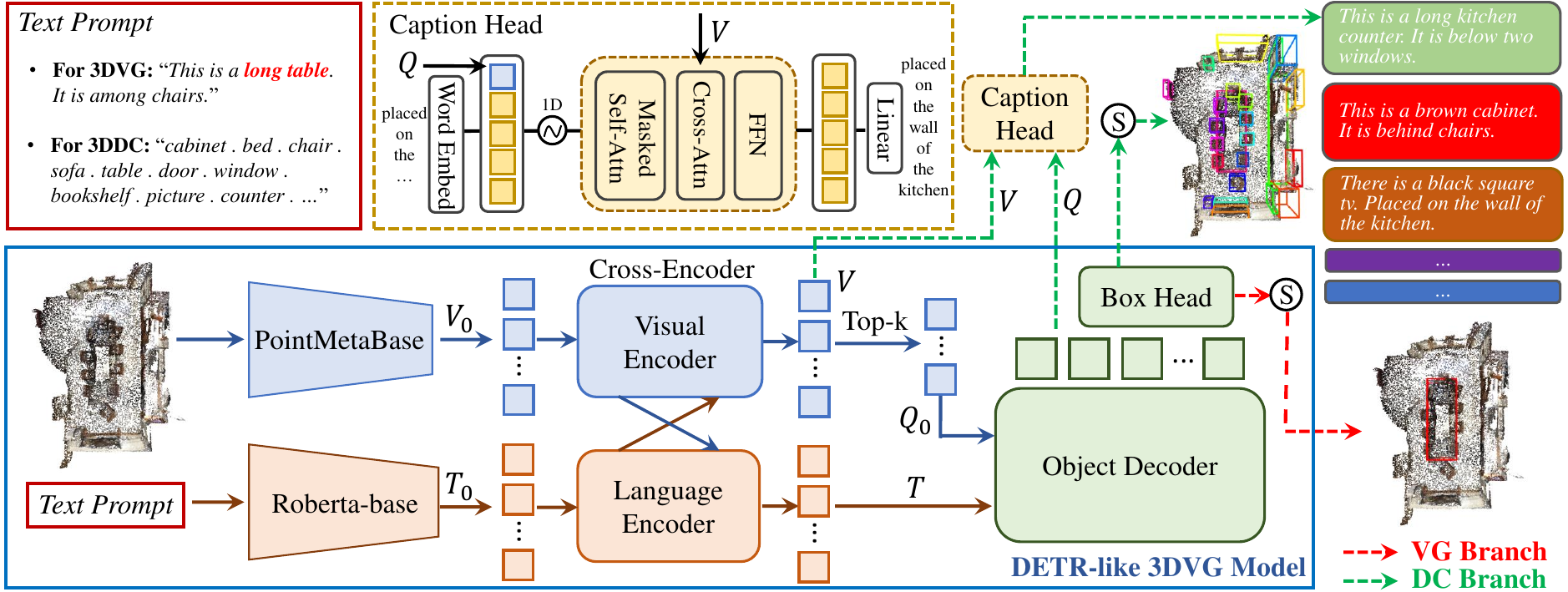}
  \caption{The framework of 3DGCTR builds upon a mature DETR-like 3DVG model (bottom in the figure) with a caption head. After obtaining the fused visual tokens $V$ and decoder output query embeddings $Q$ of each scene, the caption head uses $Q$ as caption prefix to identify the described region, and contextual features $V$ surrounding the vote query to complement with more surrounding information for more descriptive caption generation. Finally, the referring/detection boxes are selected from the candidate boxes via the referring scores. }
  \label{fig_framework} 
\end{figure*}

\section{Method}\label{method}

The key idea of 3DGCTR is to harness the relation-oriented capability of the 3DVG task and the object-oriented ability of the 3DDC task to complement each other.
The architecture of the adopted 3DVG model is illustrated in Sec. \ref{preliminary}. 
We also enhance the feature extraction ability of EDA \citep{eda} as our 3DVG model, termed EDA-PMB, which is introduced in Sec. \ref{eda_pmb}.
To develop the 3DDC capacity, a caption head is appended to the 3DVG model for caption generation, which is detailed in Sec. \ref{caption_head}. 
To naturally unify 3DDC and 3DVG tasks in DETR-like architecture, we propose a text prompt for 3DDC as an input to the 3DVG backbone, as described in Sec. \ref{caption_prompt}.  
Last, the training scheme and the inference process are illustrated in Sec. \ref{training} and Sec. \ref{inference}, respectively.

\subsection{Preliminaries: 3DVG Model}\label{preliminary}

To ensure the end-to-end training characteristics of our model, we introduce EDA \citep{eda} with the setting of single-stage as the basic 3DVG model. EDA is a typical DETR-like \citep{detr} model composed of a backbone, a cross-encoder, and an object decoder.

\textbf{Visual and Textual Backbone. }
As shown in Figure. \ref{fig_framework}, PointMetaBased \citep{pmb} produces the visual tokens 
$V_0 \in \mathbb{R}^{n \times d}$ from the input point cloud 
$P\in \mathbb{R}^{N \times 3}$, while Roberta-base \citep{roberta} processes referential text for text tokens $T_0 \in \mathbb{R}^{l \times d}$, where $n$ denotes the number of visual token, $l$ is the number of text token, $N$ denotes the point cloud size, and $d$ is the feature dimension. 

\textbf{Cross-Encoder. }
The visual tokens $V_0 $ and text tokens $T_0$ enter a dual-pathway cross-encoder alternatively stacking with self-attention, cross-attention, and FFN layers, producing thoroughly fused features $V \in \mathbb{R}^{n \times d}$ and $T \in \mathbb{R}^{l \times d}$. 
The output visual tokens $V \in \mathbb{R}^{n \times d}$ is input to the KPS module proposed by \citep{groupfree}. 
The top-k object queries $Q_0 \in \mathbb{R}^{k \times d}$ are chosen and then fed into the object decoder, caption head and the query scoring branch.  

\textbf{Object Decoder. }
The top-k object queries $Q_0$ and the text tokens $T$ enter the object decoder which consists of stacked Transformer decoder layers, producing the final query embeddings $Q \in \mathbb{R}^{k \times d}$.
On top of these object query embeddings $Q$, the decoder has two branches respectively for box regression and box-text alignment. 
The box branch dynamically updates boxes $B\in \mathbb{R}^{k \times 6}$ and provides position embeddings for query embeddings in each Transformer decoder layer. 
The alignment branch outputs the referring scores $s\in \mathbb{R}^{k}$ for the query embeddings to determine which query best matches the referent. 
Given the existing box and alignment branches, the query embeddings $Q$ in the 3DVG model exhibit inherent localization abilities. In the caption branch, referring scores are also used to filter scene objects, thus describing the scene as a set of multiple objects.

\subsection{Integrating PointMetaBase into EDA}\label{eda_pmb}

A more powerful visual backbone can better extract the attributes and relative position relationships of objects in 3D scenes, which is helpful for more fine-grained interaction with language. Hence, we replaced the PointNet++ \citep{pointnet2} in EDA \citep{eda} with PointMetaBase \citep{pmb}. 
The main contribution of PointMetaBase is to abstract the point cloud feature extraction network into a \textit{Meta Architecture} consisting of four key modules and explore the optimal design of each module. However, as the K-Nearest query is used in the module \textit{Neighbor Update}, the number of downsampled point clouds output by the backbone is not fixed, and thus can't be used as the visual tokens $V_0$ directly.
To solve this problem, we design a parameter-free visual token query module. We first use the farthest point sampling (FPS) algorithm to sample a fixed number of $n$ points from the original point cloud $P$ as the candidates $C$:
\[ C=FPS(P \in \mathbb{R}^{N \times 3}) \in \mathbb{R}^{n \times 3} \]
Then, given the output tokens $V' \in \mathbb{R}^{n' \times 3}$ from the PointMetaBase, we employ the ball query \citep{pointnet2} with a ball radius of $r$ to find $k_q$ nearest neighbor tokens in the coordinate space of $V'$ for each candidate, then assign the features of $V'$ to them using max-pooling. Finally, we can obtain the visual tokens $V_0 \in \mathbb{R}^{n \times 3}$:
\[ V_0 = MaxPool(BallQuery(V', C, k_q, r) \in \mathbb{R}^{n \times 3 \times k_q} ) \]

\subsection{Caption Head}\label{caption_head}

We argue that the key to unambiguous detailed caption generation is to obtain the relationship between the target object and its close surroundings. 
As described in Vote2Cap-DETR \citep{vote2cap}, the vote queries fail to provide adequate attribute and spatial relations. To address this issue, they proposed a caption head named Dual-Clued Captioner (DCC), which introduced the vote queries’ $k$ nearest local context token features as their local surroundings and keys for cross attention.
In our method, the top-k object queries $Q_0$ generated in the KPS module also have the same problem as the vote queries generated in Vote2Cap-DETR.
However, different from Vote2Cap-DETR, where vote queries are constructed based on spatial bias and content information for better caption generation, our KPS-based method focuses on the precise localization of key points of objects to tackle both 3DVG and 3DDC tasks simultaneously.
Therefore, we use DCC as our caption head with some modifications to fit this difference. Instead of using vote queries’ $k$ nearest for cross attention, we introduce the visual tokens $V$ that integrate the broader scene context as the object's interaction and relation information with its surroundings, leading to more contextually nuanced captions.

As shown in Figure \ref{fig_framework}, DCC is a lightweight transformer decoder-based model comprising two identical transformer decoder blocks, sinusoidal position embeddings $PE(\cdot)$, and a linear classification head.
For effective caption generation, DCC processes dual visual cues $V_c=(Q, V)$. 
Firstly, following \citep{sgt}, in captioning a proposal, the standard 'Start Of Sequences' (SOS) prefix is replaced with the query $Q$ from the described query, identifying the object in focus.
Then, for each query embedding $q \in Q$, DDC gives its corresponding dense caption. The masked self-attention and cross-attention mechanism of the transformer decodes the contextual relationship between the query and the visual scene, which can be formulated as:
\[ Atten(V_c)=CrossAtten(SelfAtten(PE(Q), mask), V) \]
Finally, the output of the cross-attention is then processed through a Feed-Forward Network (FFN) together with the linear classification head to map the processed features to the caption vocabulary:
\[ Captions=Linear(FFN(Atten(V_c))) \]

\subsection{Caption Text Prompt}\label{caption_prompt}
BUTD-DETR \citep{butd} suggests that the task of object detection is essentially a form of referential language grounding, where the utterance is simply the object's category label. Essentially, object detection can be viewed as the grounding process of detection prompts.

Concretely, utilizing the detector's library of object categories, we form prompts by stringing together the labels of objects to be identified, like ``\textit{cabinet . bed . chair . sofa . }". These concatenated labels are treated as reference text for grounding: the goal is to locate all instances of the mentioned categories in the scene, if present. Additionally, incorporating negative category labels (for which no instances exist in the scene) serves as negative training, teaching the model to avoid matching any boxes to these negative labels.

Through this approach, we can perform visual grounding and dense captioning in a single-stage manner. This integration enables end-to-end training within a single framework, fostering a synergistic relationship where each task mutually reinforces the other.

\subsection{Multi-task End-to-end Training}\label{training}

For the 3DVG task, we follow the training scheme of EDA \citep{eda}, which adopts five loss functions for each layer of the object decoder:  
box center coordinate prediction with a smooth-$L_1$ loss $\mathcal{L}_{\text {coord }}$, 
box size prediction with a smooth-$L_1$ loss $\mathcal{L}_{\text {size }}$, 
GIoU loss \citep{giouloss} $\mathcal{L}_{\text {giou }}$, 
the semantic alignment loss $\mathcal{L}_{\text {sem }}$,
and the position alignment loss $\mathcal{L}_{\text {pos }}$. 
The loss of $l$-th decoder layer is the combination of these 5 loss terms by weighted summation: 

\begin{small}
$$
\mathcal{L}_{\text {dec }}^{(l)}=\beta_1 \mathcal{L}_{\text {coord }}^{(l)}+\beta_2 \mathcal{L}_{\text {size }}^{(l)}+\beta_3 \mathcal{L}_{\text {giou }}^{(l)}+\beta_4 \mathcal{L}_{\text {sem }}^{(l)}+\beta_5 \mathcal{L}_{\text {pos }}^{(l)}.
$$
\end{small}

The losses on all decoder layers are averaged to form the total 3DVG loss: $\mathcal{L}_{\text {vg }} = \frac{1}{L} \sum_{l=1}^L \mathcal{L}_{\text {dec }}^{(l)}$.

For the 3DDC task, we follow Vote2Cap-DETR \citep{vote2cap} and apply the standard cross-entropy loss (MLE training), then fine-tune it with Self-Critical Sequence Training (SCST). We adopt the standard SCST in our model, whose reward function is CIDEr \citep{cider} score.
We termed the caption loss as $\mathcal{L}_{\text {cap }}$. During MLE training, together with the KPS loss \citep{groupfree} $\mathcal{L}_{\text {kps }}$ for the query selection, the final loss for 3DGCTR in end-to-end manner is as follows: 
$$
\mathcal{L}=\alpha_1\mathcal{L}_{\text {vg}}+\alpha_2\mathcal{L}_{\text {cap}}+\alpha_3\mathcal{L}_{\text {kps }}.
$$


\subsection{Inference}\label{inference}
During inference, given an input point cloud and a referring text, 3DGCTR outputs the object boxes $B\in \mathbb{R}^{k \times 6}$. 
Subsequently, the referent box will be selected according to the referring scores $s\in \mathbb{R}^{k}$. 
In 3DVG, the target box described in the referring text will be selected, while in 3DDC, all boxes described in the prompt are selected, and the caption head provides a dense caption for each box.

\section{Experiments}

\subsection{DataSets and Metrics}\label{datasets}
We utilized two widely recognized datasets: ScanRefer \citep{scanrefer} and Nr3D \citep{referit3d}, both of which are based on ScanNet's \citep{scannet} 3D indoor scenes. Specifically, ScanNet \citep{scannet} comprises 1,201 training and 312 validation indoor 3D scenes. The ScanRefer \citep{scanrefer} and Nr3D \citep{referit3d} datasets include 36,665 and 32,919 free-form language descriptions, respectively, referencing 7,875 and 4,664 objects within 562 and 511 training scenes. The evaluation was conducted on 9,508 and 8,584 sentences corresponding to 2,068 and 1,214 objects across 141 and 130 3D scenes, respectively.

In terms of the evaluation metric, for 3DVG, we use Acc@IoU to measure the proportion of descriptions where the predicted box and ground truth overlap with an IoU greater than 0.25 and 0.5. Descriptions are categorized into ``unique" if the object is the sole representative of its class in the scene, or ``multiple" otherwise. As for 3DDC, following \citep{vote2cap}, the evaluation metric is $m$@$k$IoU that $m$ could be any metric for natural language generation, such as CIDEr \citep{cider}, METEOR \citep{meteor}, BLEU-4 \citep{bleu}, and ROUGE-L \citep{rouge}. In practice, $k$ is set to 0.25 and 0.5.

\subsection{Implementation Details}\label{implementation_details}
We first pre-train the 3DVG model without the caption head on ScanRefer \citep{scanrefer} and Nr3D\citep{referit3d} datasets following the training settings in \citep{eda} on four NVIDIA V100 GPUs for 4 days.
During MLE training, we load the pre-trained 3DVG weight and jointly train the whole network for another 30 epochs in both VG and DC tasks for 12 hours.
To prevent overfitting, the initial learning rates of the 3DVG model and caption head are set to 2e-6 and 2e-4 for ScanRefer \citep{scanrefer} (1e-4 and 1e-6 for Nr3D \citep{referit3d}), respectively.
We apply learning rate decay at epochs 10 and 20 with a rate of 0.1. As for the input type, we use XYZ coordinates and RGB values as the input, and the number of visual tokens $V$ and query embeddings $Q$ are empirically set to 1024 and 256, respectively. 
The number of decoder layers $L$ is set to 6.
The balancing factors are set default as $\alpha_1$ = 1.0 / ($L$ + 1), $\alpha_2$ = 5.0, $\alpha_3$ = 8.0, $\beta_1$ = 5.0, $\beta_2$ = 1.0, $\beta_3$ = 1.0, $\beta_4$ = 0.5 and $\beta_5$ = 0.5 for the ScanRefer dataset. For the Nr3D dataset, $\beta_4$ and $\beta_5$ are adjusted to be 1.0 and 1.0, respectively.
During SCST training, we tune the caption head with a batch size of 8 and freeze the rest of the modules in single-GPU for 8 hours. We train 400 epochs with a fixed learning rate of 5e-6 and decay learning rate at epochs 100 and 200 with a rate of 0.1.

\begin{table*}[h]
  \centering
\setlength{\tabcolsep}{4.5mm}  
\renewcommand{\arraystretch}{1.05} 
\caption{The 3D referring expression comprehension results on ScanRefer in terms of Acc@0.25IoU and Acc@0.5IoU. Note that what is shown in the table is the optimal accuracy of each method as reported in its paper. Our 3DGCTR surpasses existing state-of-the-art works by a significant margin. EDA-PMB refers to the replacement of single-stage EDA's \citep{eda} visual backbone from Pointnet++ \citep{pointnet2} to PointMetaBase \citep{pmb}. }
\label{tab_3dvg}
\scalebox{0.9}[0.9]{\begin{tabular}{l|cc|cc|cc}
\hline
\multicolumn{1}{c|}{} & \multicolumn{2}{c|}{Unique(19\%)} & \multicolumn{2}{c|}{Multiple(81\%)} & \multicolumn{2}{c}{Overall} \\
\multicolumn{1}{c|}{\multirow{-2}{*}{Method}}   & 0.25   & 0.5            & 0.25    & 0.5   & 0.25  & 0.5   \\ \hline
ScanRefer \citep{scanrefer} & 67.64 & 49.19 & 32.06 & 21.26 & 38.97 & 26.10 \\
TGNN \citep{tgnn}           & 68.61 & 56.80 & 29.84 & 23.18 & 37.37 & 29.70 \\
3DJCG \citep{3djcg}         & 78.75 & 61.30 & 40.13 & 30.08 & 47.62 & 36.14 \\
D3Net \citep{d3net}         & -     & 70.35 & -     & 30.50 & -     & 37.87 \\
UniT3D \citep{unit3d}       & 82.75 & 73.14 & 36.36 & 31.05 & 45.27     & 39.14 \\
3DSPS \citep{3dsps}         & 84.12 & 66.72 & 40.32 & 29.82 & 48.82 & 36.98 \\
BUTD-DETR \citep{butd}           & 82.88 & 64.98 & 44.73 & 33.97 & 50.42 & 38.60 \\
EDA \citep{eda}             & 85.76 & 68.57 & 49.13 & 37.64 & 54.59 & 
42.26 \\
\rowcolor{gray!10} \textbf{EDA-PMB (our backbone)}                   & \textbf{88.22} & \textbf{73.40} & 51.76 & 40.82 & 57.45 & 45.91 \\
\rowcolor{gray!10} \textbf{3DGCTR (ours)} & 88.01 & 73.13 & \textbf{52.15} & \textbf{41.31} & \textbf{57.75} & \textbf{46.28} \\ \hline
\end{tabular}
}
\end{table*}

\subsection{Quantitative Comparison}\label{quantitative_comparison}

In this subsection, we perform a qualitative comparison on ScanRefer \citep{scanrefer} and Nr3D \citep{referit3d} datasets. 

\textbf{3DVG.}
To evaluate the 3DVG performance, we compare 3DGCTR with several existing 3DVG works in the 3D Referring Expression Comprehension (3DREC) task on ScanRefer \citep{scanrefer}, involving the state-of-the-art DETR-like methods EDA \citep{eda} and BUTD-DETR \citep{butd}, among others.
As evidenced in Table. \ref{tab_3dvg}, 3DGCTR consistently exhibits superior performance over these prevailing methods in both two-stage and single-stage settings. 
This evidence strongly suggests that the newly incorporated caption head has indeed augmented the performance of the 3DVG task.
Specifically, by co-training the two tasks, 3DGCTR gets a 0.3\% Acc@0.5IoU improvement in overall performance compared to the pre-trained 3DVG backbone EDA-PMB we used. 
Although there is a slight decrease in the ``unique" setting, our 3DGCTR performs better in more complex and ``multiple" scenarios (81\% of the total testing set), suggesting that the dense caption task can lead to stronger semantic alignment for visual grounding task.
The commendable results can be largely attributed to the well-constructed design of our unified 3DVG framework.
We also conduct experiments on Nr3D \citep{referit3d}, setting the state-of-the-art performance as shown in Table. \ref{tab_visual_nr3d}.

\textbf{3DDC. }
In 3DDC performance, among the evaluated methods, most employ the standard VoteNet \citep{votenet} detector, except D3Net \citep{d3net} and 3DJCG \citep{3djcg}.
The state-of-the-art method Vote2Cap-DETR \citep{vote2cap} is the first attempt to jointly train 3D Detection and 3D Dense Caption tasks in a single-stage manner, which gains state-of-the-art performance. Table. \ref{tab_3ddc} reports comparisons on ScanRefer \citep{scanrefer} dataset. 
Our model prioritizes enhancing caption accuracy and key content extraction, such as spatial positioning and object attributes, potentially affecting sentence fluidity which leads to a slight performance drop in M and R indicators that focus on the fluency.
However, our model excels in the C metric is specially designed for visual description tasks, which offers superior referential value. Specifically, our 3DGCTR achieves 66.11\% C@0.5 while Vote2Cap-DETR. \citep{vote2cap} achieves 61.81\% (4.30\% C@0.5↑). Additionally, under SCST, our 3DGCTR outperforms the state-of-the-art method C@0.5 (0.18\% C@0.5↑).
As shown in Table. \ref{tab_caption_nr3d}, a large boost on Nr3D \citep{referit3d} dataset suggests that our pipeline has a greater advantage in dealing with scenes with more complex descriptions.

\begin{table*}[t]
  \centering
\caption{The 3D referring expression comprehension results on ScanRefer in terms of CIDEr (\textbf{C}), METEOR (\textbf{M}), BLEU-4 (\textbf{B-4}) and ROUGE-L (\textbf{R}), with the setting of $m$@0.25IoU and $m$@0.5IoU in both four metrics. We compare 3DGCTR with all published  3D dense caption methods on the ScanRefer dataset, which demonstrates that 3DGCRE achieves new state-of-the-art under both MLE and SCST training.}
\label{tab_3ddc}
\scalebox{0.9}[0.9]{\setlength{\tabcolsep}{3mm}  
\renewcommand{\arraystretch}{1.05} 
\begin{tabular}{l|cccc|cccc}
\hline
\multicolumn{1}{c|}{} & \multicolumn{4}{c|}{IoU=0.25} & \multicolumn{4}{c}{IoU=0.5} \\
\multicolumn{1}{c|}{\multirow{-2}{*}{Method}} & C & B-4 & M & R & C & B-4 & M & R        \\ \hline
\multicolumn{7}{l}{\textit{MLE training}} \\ 
\hline
Scan3Cap \citep{scan2cap} & 56.82 & 34.18 & 26.29 & 55.27 & 39.08 & 23.32 & 21.97 & 44.78 \\
MORE \citep{more}         & 62.91 & 36.25 & 26.75 & 56.33 & 40.94 & 22.93 & 21.66 & 44.42 \\
SpaCap3d \citep{spacap}   & 63.30 & 36.46 & 26.71 & 55.71 & 44.02 & 25.26 & 22.33 & 45.36 \\
3DJCG \citep{3djcg}       & 64.70 & 40.17 & 27.66 & 59.23 & 49.48 & 31.03 & 24.22 & 50.80 \\
D3Net \citep{d3net}       & - & - & - & - & 46.07 & 30.29 & 24.35 & 51.67 \\
UniT3D \citep{unit3d}       & - & - & - & - & 46.69 & 27.22 & 21.91 & 45.98 \\
Vote2Cap-DETR \citep{vote2cap} & 71.45 & 39.34 & 28.25 & 59.33 & 61.81 & 34.46 & \textbf{26.22} & \textbf{54.40} \\
\rowcolor{gray!10} \textbf{3DGCTR (ours)} & \textbf{84.87} & \textbf{44.58} & \textbf{29.68} & \textbf{63.24} & \textbf{66.11} & \textbf{35.85} & 26.12 & 54.29\\ \hline
 \multicolumn{7}{l}{\textit{SCST training}} \\ 
\hline
$\mathcal{X}$-Trans2Cap \citep{Xtrans2cap}   & 61.83 & 35.65 & 26.61 & 54.70 & 43.87 & 25.05 & 22.46 & 45.28 \\
Scan3Cap \citep{scan2cap}  & - & - & - & - & 48.38 & 26.09 & 22.15 & 44.74
 \\
D3Net \citep{d3net}         & - & - & - & - & 62.64 & 35.68 & 25.72 & 53.90
 \\
Vote2Cap-DETR \citep{vote2cap}  & 84.15 & 42.51 & 28.47 & 59.26 & 73.77 & 38.21 & \textbf{26.64} & 54.71 \\
\rowcolor{gray!10} \textbf{3DGCTR (ours)} & \textbf{95.96} & \textbf{48.13} & \textbf{29.91} & \textbf{64.23} & \textbf{73.95} & \textbf{38.71} & 26.29 & \textbf{56.27} \\ \hline
\end{tabular}
}
\end{table*}


\begin{table}[ht]
\centering
\resizebox{1.0\textwidth}{!}{ 
    \begin{minipage}[c]{0.5\linewidth}
        \centering
        \renewcommand{\arraystretch}{1.1}
        \caption{The 3DVG performance on Nr3D datasets by Acc@0.25IoU as the metric.  }
        \label{tab_visual_nr3d}
        \setlength{\tabcolsep}{4.0mm}  
        \begin{tabular}{l|cc}
        \hline
        \multicolumn{1}{c|}{Method}      & Acc@0.25IoU  \\ \hline
        ReferIt3D \citep{referit3d}   & 35.6  \\
        TGNN \citep{tgnn}       & 37.3 \\
        3DSPS \citep{3dsps}     & 51.5   \\
        BUTD \citep{butd}     & 49.1  \\
        EDA \citep{eda}        & 52.1   \\
        \rowcolor{gray!10}\textbf{3DGCTR (ours)}  & \textbf{52.4}   \\ \hline
        \end{tabular}
    \end{minipage}
    \hspace{0.5cm}
    \begin{minipage}[c]{0.54\linewidth}
        \centering
        \renewcommand{\arraystretch}{1.1}
        \caption{The 3DDC performance on Nr3D datasets by CIDer@0.5IoU under MLE and SCST training. }
        \label{tab_caption_nr3d}
        \setlength{\tabcolsep}{3.8mm}  
        \begin{tabular}{l|cc}
        \hline
        \multicolumn{1}{c|}{Method}    & MLE & SCST\\ \hline
        SpaCap3d\citep{spacap}  & 33.71 & -  \\
        $\mathcal{X}$-Trans2Cap\citep{Xtrans2cap}  & - & 33.62  \\
        D3Net\citep{d3net}    & 33.85 & 38.42 \\
        3DJCG\citep{3djcg}     & 38.06 & - \\
        Vote2Cap-DERT\citep{vote2cap}  & 43.84 & 45.53   \\
        \rowcolor{gray!10} \textbf{3DCGTR (ours)}  & \textbf{49.89} & \textbf{52.07}  \\ \hline
        \end{tabular}
    \end{minipage}
    }
\end{table}

\subsection{Ablation Study}

The experiments in this subsection are conducted on Scanrefer using MLE training, and metrics adopted for 3DVG and 3DDC performance are Acc@0.25IoU and CIDer@0.5IoU, respectively.

\textbf{Influence of 3DVG model.}
We replace the pre-trained 3DVG model with EDA \citep{eda} in both single- and two-stage settings. As illustrated in Table. \ref{tab_backbone}, under the joint training setting, various 3DVG backbones achieve enhanced performance in 3DVG tasks (Acc@0.25IoU). Meanwhile, when using the state-of-the-art 3DVG model \citep{eda} as our component, we obtain the state-of-the-art 3DDC performance compared to Vote2Cap-DETR \citep{vote2cap} (3.29\% C@0.5↑), which demonstrate that our ``3DDC builds upon 3DVG model" rethinking pipeline offers greater advantages.


\textbf{Joint Training Scheme.}
We explore the impacts of different joint training schemes on task performance.
After pre-training the 3DVG model without the caption head, if only dense caption data is used for training, the training scheme can be divided into the following three types:
1) only 3DDC: training all components with the same learning rate, 
2) only 3DDC (two lr): training all components with different learning rates ($2e-4$ and $2e-6$),
3) only 3DDC (frozen vg): freezing 3DVG components and training 3DDC components.
If both data to jointly train 3DDC and 3DVG tasks, the training scheme can be divided into the following two types:
4) joint training: jointly training 3DDC and 3DVG with the same learning rate, 
5) joint training (two lr): jointly training 3DDC and 3DVG with different learning rates ($2e-4$ and $2e-6$),
As displayed in Table. \ref{tab_traing}, 
The initial three schemes enhance 3DDC but not 3DVG. The first scheme's uniform learning rate degrades 3DVG by biasing the model towards DC data, affecting batch normalization statistics. The second and third schemes prevent this by freezing or reducing the 3DVG learning rate, thus maintaining 3DVG accuracy and boosting 3DDC. Conversely, in joint training setting, the fourth scheme's equal learning rates for both tasks lead to 3DVG overfitting and compromised 3DDC. The fifth scheme's differential learning rates yield the best 3DDC and improved 3DVG outcomes. Joint training under the fifth scheme shows a 0.3\% 3DVG and 1.27\% 3DDC performance increase, indicating that a strategic training regimen can effectively co-train 3DDC and 3DVG, mutually enhancing both tasks.

\begin{table}[ht]
\resizebox{1.0\textwidth}{!}{ 
\centering
\begin{minipage}[c]{0.5\linewidth} 
\centering
\caption{Performance on Scanrefer affected by different 3DVG backbone during MLE training. The expressions EDA$^{s}$ and EDA$^{t}$ denote the application of EDA \citep{eda} in single- and two-stage settings, respectively. The acronyms 3DGCTR-EDA$^{s}$ and 3DGCTR-EDA$^{t}$ are used to indicate the utilization of EDA$^{s}$ and EDA$^{t}$ as the 3DVG backbone.}
\label{tab_backbone}
\setlength{\tabcolsep}{2.0mm}  
\begin{tabular}{l|cc}
\hline
\multicolumn{1}{c|}{Method} & 3DDC   & 3DVG          \\ \hline
Vote2Cap-DETR \citep{vote2cap}         & 61.81       & - \\  \hline
EDA$^{s}$ \citep{eda}         & -          &  53.83         \\ 
3DGCTR-EDA$^{s}$      &   58.85  &   54.37\small\color{red}\textbf{(+0.54\%)}           \\ \hline
EDA$^{t}$ \citep{eda}         & -          &  54.59         \\ 
3DGCTR-EDA$^{t}$      &  65.10   &   55.09\small\color{red}\textbf{(+0.50\%)}           \\ \hline
EDA-PMB     & -        & 57.45              \\ 
3DGCTR     &  66.11        &  57.75\small\color{red}\textbf{(+0.30\%)}       \\ \hline 
\end{tabular}
\end{minipage}
\hspace{0.5cm} 
\begin{minipage}[c]{0.5\linewidth}
\centering
\renewcommand{\arraystretch}{1.2}
\caption{Performance on Scanrefer affected by different training schemes during MLE training. 'two lr' means jointly training 3DDC and 3DVG components with different learning rates, and 'frozen vg' means only training 3DDC components while freezing the 3DVG backbone. }
\label{tab_traing}
\setlength{\tabcolsep}{2.2mm}  
\begin{tabular}{l|cc}
\hline
\multicolumn{1}{c|}{Training Scheme} & 3DDC   & 3DVG          \\ \hline
EDA-PMB          & -              & 57.45       \\
Vote2Cap-DETR \citep{vote2cap}   & 61.81          & -          \\  \hline
1) only 3DDC                & 62.89          & 44.84              \\
2) only 3DDC (two lr)       & 64.84          & 57.46              \\ 
3) only 3DDC (frozen vg)    & 63.83          & 57.45         \\ \hline
4) joint training           & 60.17          & 55.85          \\
5) joint training (two lr)  & \textbf{66.11} & \textbf{57.75}      \\
 \hline
\end{tabular}

\end{minipage}
}
\end{table}


\subsection{Qualitative Comparison}
\begin{figure*}[h!]
  \centering 
  \includegraphics[width=1.0\linewidth]{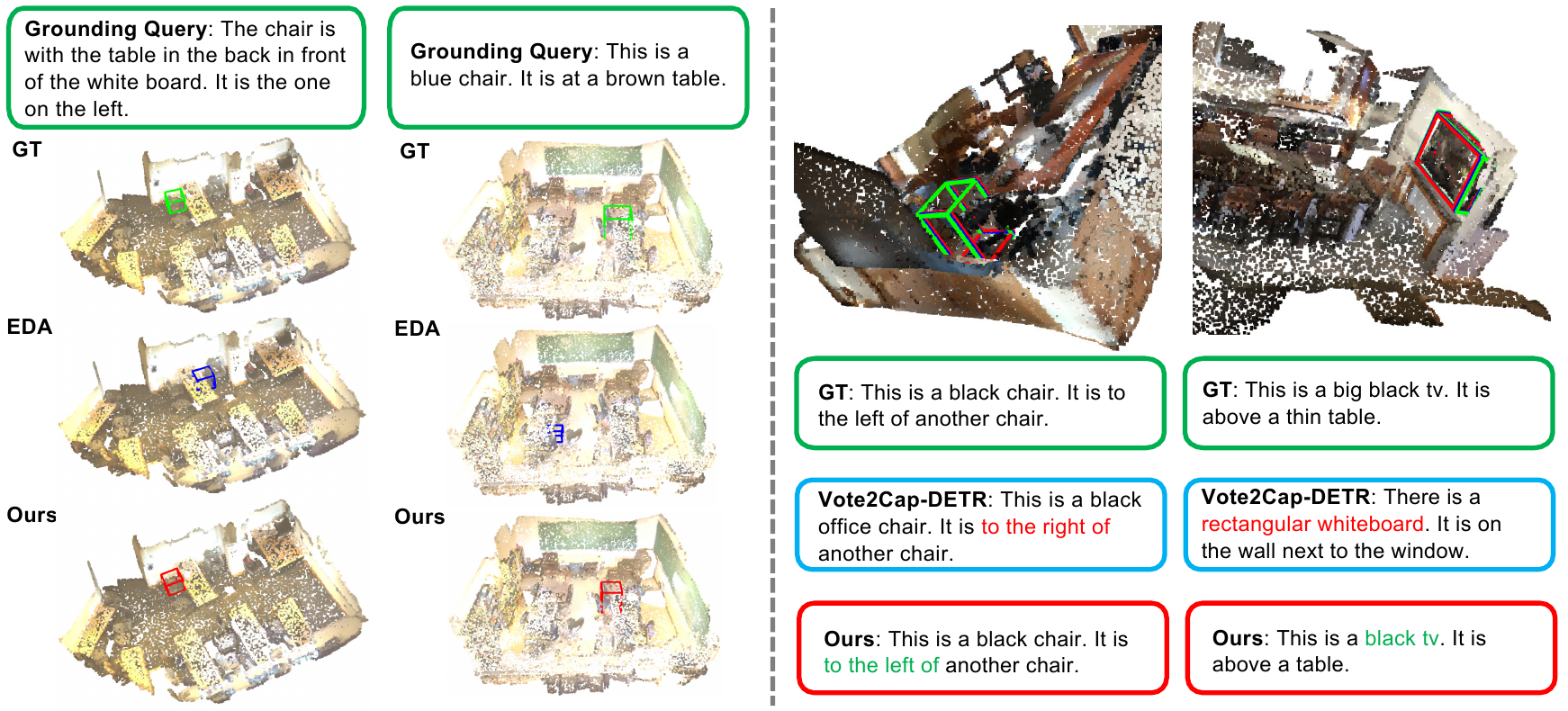}
  \caption{Qualitative Comparisons. We compare qualitative results with two state-of-the-art methods in 3DVG (left part in the figure) and 3DDC (right part in the figure) tasks, EDA \citep{eda} and Vote2Cap-DETR \citep{vote2cap}. We mark correct attribute words in \textcolor{green}{green} and wrong descriptions in \textcolor{red}{red}. Our method produces right bounding boxes close to ground truth annotations and produces accurate descriptions of object attributes, classes and spatial relationships.}
  \label{fig_vis} 
\end{figure*}

As for the 3DVG task, we compare qualitative results to the state-of-the-art model EDA \citep{eda} in the left part of Figure \ref{fig_vis}. Our method produces the right bounding boxes in some difficult samples, which indicates that by jointly training the 3DDC and 3DVG tasks, the 3DVG task benefits from the 3DDC task's comprehensive characterization of object attributes.
As for the 3DDC task, we compare qualitative results to the state-of-the-art model Vote2Cap-DETR \citep{vote2cap} in the right part of Figure \ref{fig_vis}. Our method can produce accurate descriptions of object attributes, classes, and spatial relationships. This suggests the 3DDC task benefits from the 3DVG task's grasp of the relationships between objects in the scene to generate more accurate caption information. 
Through qualitative comparison, we can more intuitively recognize the advantages of an end-to-end training scheme compared to the two-stage approach.

\section{Limitations \& Conclusions} \label{conclusions}
\textbf{Limitations.}
The main limitation is scene sampling. Small objects with few sampled points may lead to a compromised understanding of detailed features. Designed for scenes with fixed limited sampled points, the model performs well in most indoor settings but may struggle with larger scenes.
\textbf{Conclusions.}
We introduce a significant shift in 3DVG and 3DDC integration using the proposed 3DGCTR. By rethinking prompt-localization in 3DVG to enhance 3DDC, this approach moves away from traditional two-stage, detector-dependent methods towards a more effective, unified strategy. The DETR-like framework enables end-to-end training, utilizing shared modules for multitasking. 3DGCTR's exceptional performance on ScanRefer and its ability to mutually enhance 3DVG and 3DDC tasks underline its transformative impact on 3D scene understanding.


\bibliographystyle{abbrvnat}
\bibliography{Styles/neurips_2024}

\end{document}